\def\year{2021}\relax
\documentclass[letterpaper]{article} 
\usepackage{aaai21}  
\usepackage{times}  
\usepackage{helvet} 
\usepackage{courier}  
\usepackage[hyphens]{url}  
\usepackage{graphicx} 
\usepackage{natbib}  
\usepackage{caption} 
\usepackage{multirow}
\usepackage{amssymb}
\usepackage{amsmath}
\usepackage[linesnumbered,ruled,vlined]{algorithm2e}
\usepackage{diagbox}
\usepackage{pgfplots}
\usepackage[switch]{lineno}
\SetKwInput{KwInput}{Input}    
\SetKwInput{KwOutput}{Output}  
\title{Sequential End-to-end Network for Efficient Person Search}
\author{

    Zhengjia Li\textsuperscript{\rm 1,2},
    Duoqian Miao\textsuperscript{\rm 1,2}\thanks{Corresponding author.}
}
\affiliations{

    \textsuperscript{\rm 1}Department of Computer Science and Technology, Tongji University, Shanghai 201804, China\\
    \textsuperscript{\rm 2}Key Laboratory of Embedded System and Service Computing, Ministry of Education, Shanghai 201804, China\\

    zjli1997@tongji.edu.cn,
    dqmiao@tongji.edu.cn

}

\begin{document}

\maketitle

\begin{abstract}
    Person search aims at jointly solving Person Detection and Person Re-identification (re-ID). Existing works have designed end-to-end networks based on Faster R-CNN. However, due to the parallel structure of Faster R-CNN, the extracted features come from the low-quality proposals generated by the Region Proposal Network, rather than the detected high-quality bounding boxes. Person search is a fine-grained task and such inferior features will significantly reduce re-ID performance. To address this issue, we propose a Sequential End-to-end Network (SeqNet) to extract superior features. In SeqNet, detection and re-ID are considered as a progressive process and tackled with two sub-networks sequentially. In addition, we design a robust Context Bipartite Graph Matching (CBGM) algorithm to effectively employ context information as an important complementary cue for person matching. Extensive experiments on two widely used person search benchmarks, CUHK-SYSU and PRW, have shown that our method achieves state-of-the-art results.
    Also, our model runs at 11.5 fps on a single GPU and can be integrated into the existing end-to-end framework easily.
\end{abstract}

\section{Introduction}
\noindent Pedestrian detection \cite{rcnn,fast-rcnn,faster-rcnn} aims at detecting the bounding boxes (BBoxes) of all people in the image. Person re-identification (re-ID) \cite{yang2017unsupervised,zhao2017consistent,wang2019spatial,fu2019horizontal,hao2019hsme,zhao2020deep} is used to match the interested person with hand-cropped person images. Although these two fields are widely studied in recent years, they can not be directly applied to real-world applications due to their limited functionality. To close the gap, Xu et al. introduce person search task which aims at locating a target person in the scene image \cite{first-ps}. Person search can be seen as a combination of pedestrian detection and person re-ID. It has broad application prospects in video surveillance, finding lost children, and self-service supermarket, \textit{etc}.

\begin{figure}[t]
    \centering
    \includegraphics[width=0.8\columnwidth]{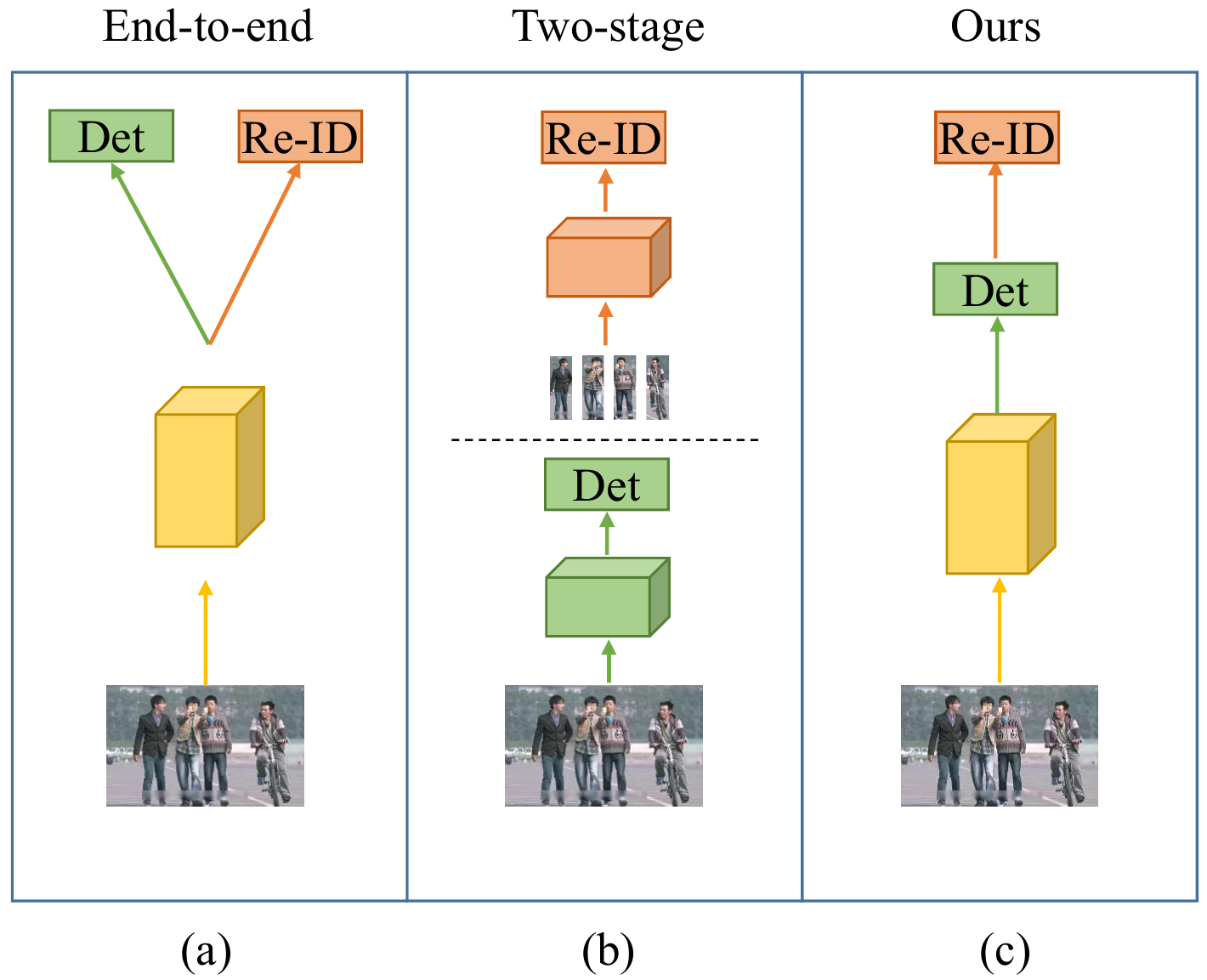}
    \caption{Comparison of three methods for person search. (a). Existing end-to-end framework. (b). Existing two-stage framework. (c). Ours.}
    \label{three_frameworks}
\end{figure}

As illustrated in Figure \ref{three_frameworks}, existing works divide the task into generating BBoxes of all people in the image and person re-ID. They either tackle the problem separately with two independent models (\textit{two-stage} methods) or jointly with a multi-task model (\textit{end-to-end} methods).

For end-to-end methods \cite{oim,ian,qeeps}, they design a multi-task framework based on Faster R-CNN \cite{faster-rcnn}. A Region Proposal Network (RPN) is built to generate region proposals, which are then fed into the subsequent parallel detection and re-ID branches. However, these features extracted by the network come from low-quality proposals rather than detected accurate BBoxes. Although these inferior features have little impact on the coarse-grained classification task, they will significantly reduce the performance of the fine-grained re-ID task. This problem is caused by the parallel structure of Faster R-CNN. Because detection and re-ID are processed at the same time, the accurate BBoxes are not available before extracting re-ID features. For two-stage methods, there is no such problem, because detection and re-ID are tackled sequentially with two separate models. However, they are time-consuming and resource-consuming.

Motivated by the above observations, we propose a Sequential End-to-end Network (SeqNet) illustrated in Figure \ref{three_frameworks} (c) to extract high-quality features. Specifically, detection and re-ID share the stem representations, but solved with two head networks sequentially. Compared with baseline, our model employs an extra Faster R-CNN head as an enhanced RPN to provide high-quality BBoxes. Then an unmodified baseline head is used to extract the discriminative features of these BBoxes. At test time, the non-maximum suppression (NMS) is applied to remove redundant BBoxes before re-ID stage for efficiency. Moreover, to improve the classification ability of baseline head for high Intersection over Union (IoU) samples, we adopt the more reliable classification result of detection head. In general, our SeqNet not only inherits the sequential process of two-stage methods, which can provide accurate BBoxes for re-ID stage, but also retains the end-to-end training fashion and efficiency.

\begin{figure}[t]
    \centering
    \includegraphics[width=0.8\columnwidth]{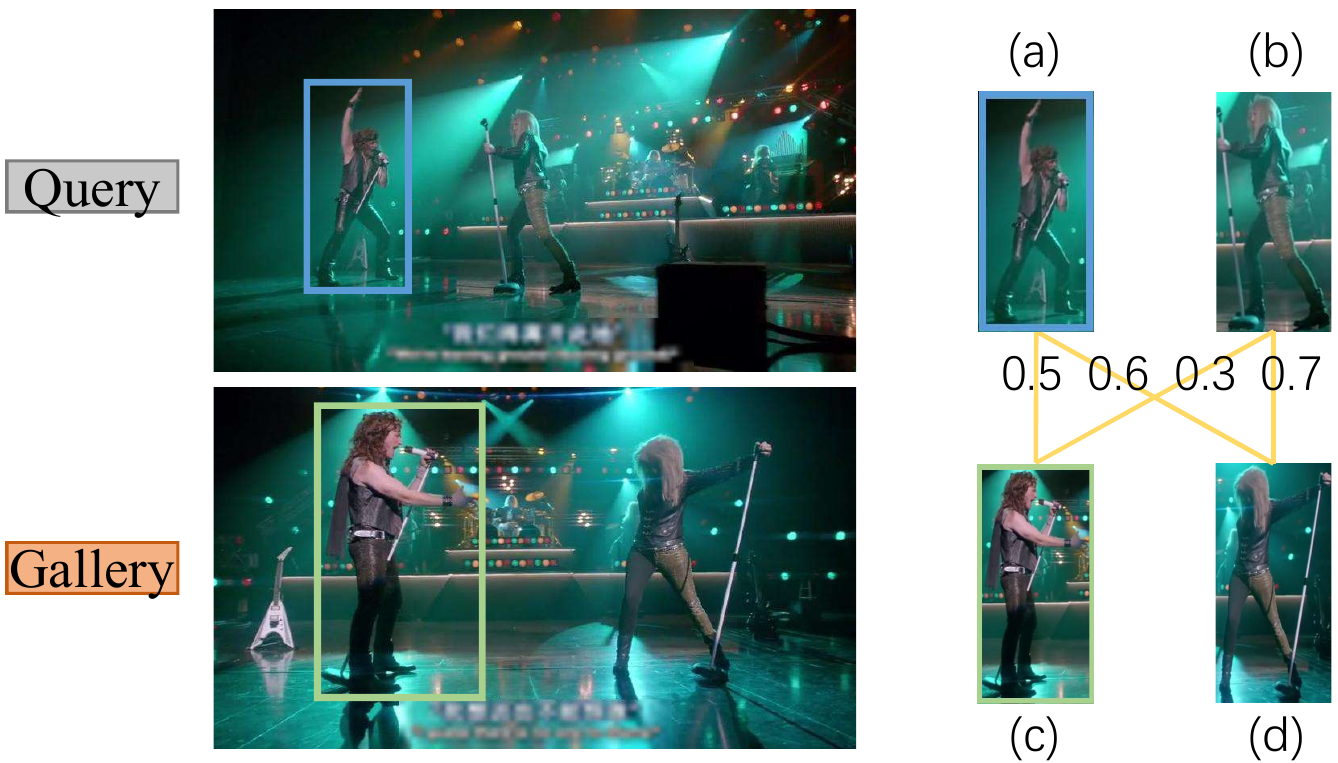}
    \caption{An example of matching query image and gallery image. Blue box denotes the query person and green box denotes the ground truth. The number attached to each yellow line represents the similarity between the two people connected by the line.}
    \label{matching_example}
\end{figure}

Another challenge for person search is how to utilize context information to perform more robust matching. As shown in Figure \ref{matching_example}, given a query person (a), (c) is the corresponding ground truth, but the (d) with largest similarity (0.6) will be mistakenly predicted as top-1 result. If context information (b) is taken into consideration, to maximize the total similarity, the optimal matching should be $(a)\leftrightarrow(c), (b)\leftrightarrow(d)$. In this way, the wrong prediction (d) can be revised to (c). Inspired by this, we design a Context Bipartite Graph Matching (CBGM) algorithm to exploit context information as a complement to individual feature. Specifically, we treat all people in the query image and each gallery image as two sets of vertices respectively. A complete bipartite graph is built upon the two sets of vertices, and the weight of each edge is the similarity between corresponding vertices calculated by the person search network. Then the Kuhn-Munkres (K-M) algorithm \cite{km-k,km-m} is exploited to discovery the optimal matching with maximum weight. In this matching, the person connected with the querier is taken as top-1 result.

The contributions of this paper are three-fold:

\begin{itemize}
    \item We notice that the performance of previous end-to-end framework is limited by inferior features and formulate a Sequential End-to-end Network (SeqNet) to refine them.
    \item To make full use of context information, we propose a Context Bipartite Graph Matching algorithm to perform more robust matching.
    \item Our method outperforms all other state-of-the-art ones on the two widely used benchmarks CUHK-SYSU \cite{oim} and PRW \cite{prw}. Moreover, our method can be integrated into the existing end-to-end framework easily.
\end{itemize}

\section{Related Work}
\subsection{Person Search}
Person search has raised a lot of interest in computer vision community since the publication of two large scale datasets, CUHK-SYSU \cite{oim} and PRW \cite{prw}. It's a straightforward solution to tackle the problem with a pedestrian detector and a re-ID descriptor sequentially. Zheng et al. make a systematic evaluation on various detectors and descriptors, and propose a re-weighting algorithm adjusting the matching similarity to suppress the false positive detections \cite{prw}. Lan, Zhu, and Gong point out the performance of person search is limited by the multi-scale matching, and formulates a Cross-Level Semantic Alignment (CLSA) method capable of learning more discriminative identity representations \cite{clsa}. Chen et al. first reveal the inherent optimization conflict between the pedestrian detection and person re-ID, and present a Mask-Guided Two-Stream (MGTS) method to eliminate the conflict \cite{mgts}. Han et al. introduce a RoI transform layer to jointly optimize the detection and re-ID models \cite{reid-driven}. Wang et al. notice the consistency requirements between the two subtasks in person search, and adopt a Task-Consistent Two-Stage (TCTS) framework to solve the inconsistency existing in previous works \cite{tcts}. Dong et al. propose a Instance Guided Proposal Network (IGPN) to reduce the number of proposals to relieve the burden of re-ID \cite{igpn}.

Besides the two-stage framework, the faster and simpler end-to-end methods based on Faster R-CNN are also popular. Xiao et al. design the first end-to-end person search network, which is trained with standard Faster R-CNN losses and their proposed Online Instance Matching (OIM) loss \cite{oim}. Xiao et al. introduce center loss to increase the intra-class compactness of feature representations \cite{ian}. Instead of generating BBoxes for all people in the image, Liu et al. propose to recursively shrinking the search area under the guidance of the query \cite{npsm}. Chang et al. adopt a similar idea and first introduce the deep reinforcement learning into person search framework \cite{rcaa}. Yan et al. build a graph model to exploit context information as an complementary cue for person matching \cite{context}. Munjal et al. propose a query-guided region proposal network (QRPN) to produce query-relevant proposals, and a query-guided similarity subnetwork (QSimNet) to learn a query-guided re-ID score \cite{qeeps}. Chen et al. propose a Hierarchical Online Instance Matching (HOIM) loss which exploits the hierarchical relationship between detection and re-ID to guide the feature learning of their network \cite{hoim}. Dong et al. design a Bi-directional Interaction Network (BINet) to remove rebundent context information outside BBoxes \cite{binet}. To reconcile the contradictory goals of the two subtasks, Chen et al. present a novel approach called Norm-Aware Embedding (NAE) to disentangle the person embedding into norm and angle for detection and re-ID respectively \cite{nae}.

\subsection{Multi-stage Faster R-CNN}
Some researchers extend Faster R-CNN to a multi-stage fashion. Gidaris and Komodakis propose a post-processing step that the network iterate several times in inference stage to achieve better localization performance \cite{iterative-bbox1,iterative-bbox2}. Cai and Vasconcelos design a cascade framework containing a sequence of detectors trained with increasing IoU thresholds to be sequentially more selective against close false positives \cite{cascade-rcnn}. Inspired by them, our model is designed as a multi-stage framework to introduce the sequential process into end-to-end person search network.

\section{Method}
In this section, we first revisit the end-to-end person search network, then discuss its shortcoming. Next, we describe our proposed Sequential End-to-end Network (SeqNet). Finally, we formulate a Context Bipartite Graph Matching (CBGM) algorithm to utilize the context information.

\begin{figure*}[t]
    \centering
    \includegraphics[width=0.8\textwidth]{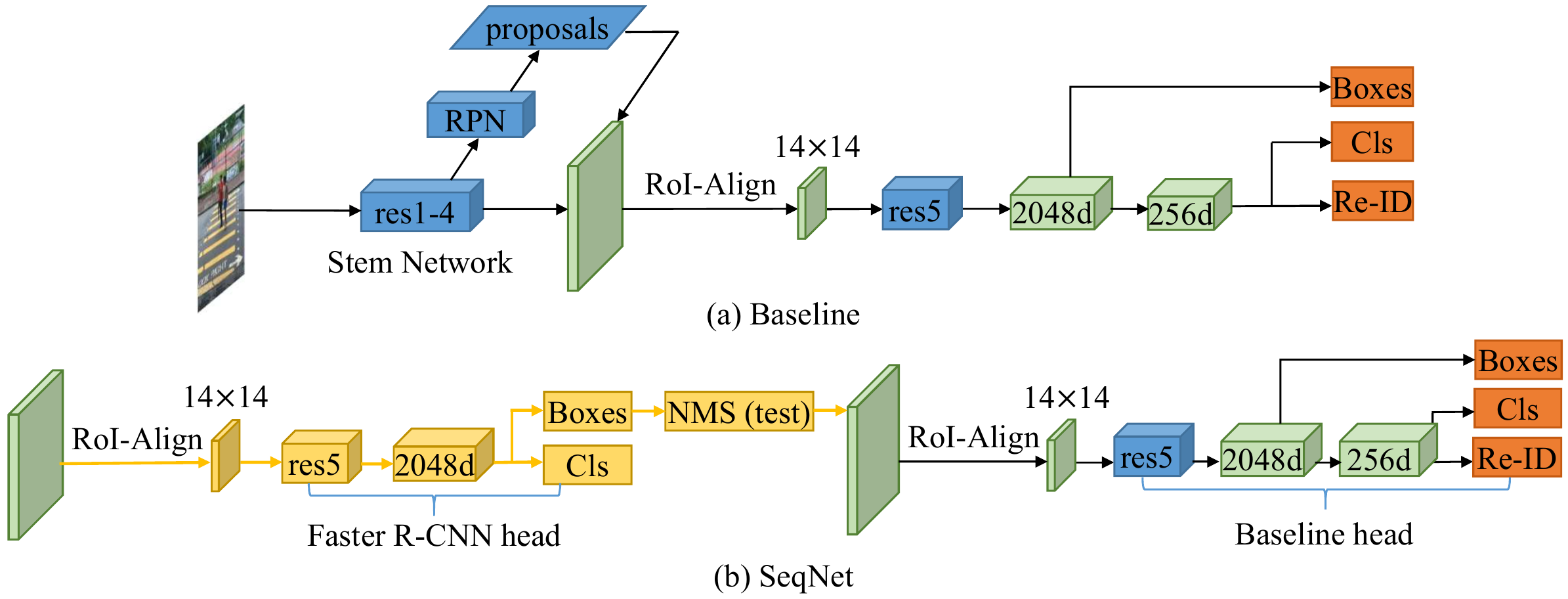}
    \caption{(a). Baseline (b). Our Sequential End-to-end Network, in which yellow parts are modifications and NMS only be applied in inference stage. The structure before RoI-Align is the same as baseline, so it is not shown here for simplification.}
    \label{net_arch}
\end{figure*}

\subsection{End-to-end Network for Person Search}
We take the multi-task network NAE \cite{nae} as our baseline. The overview of this baseline is illustrated in Figure \ref{net_arch} (a). It adopt ResNet50 \cite{resnet} as the backbone network. Specifically, res1$\sim$res4 are taken as the stem network to extract the 1024-channel stem feature maps of the image. A Region Proposal Network (RPN) is built upon these feature maps to generate region proposals. After NMS, we keep 128 proposals, and exploit RoI-Align to pool a $1024\times14\times14$ region for each of them. Next these regions are fed into res5 to extract 2048-dim features, which are then mapped to 256-dim. It uses these 2048-dim features to calculate regressors and 256-dim features to perform classification and re-ID tasks. The Norm-Aware-Embedding is designed to supervise the classification and re-ID branches and the Smooth-L$_1$-Loss \cite{fast-rcnn} is adopted to supervise the regression branch.

\subsection{Problems of the End-to-end Framework}
\begin{table}[t]
    \centering
    \resizebox{0.8\columnwidth}{!}
    {
    \begin{tabular}{|c|cc|cc|}
        \hline
        \multirow{2}{*}{\textbf{Method}} & \multicolumn{2}{c|}{\textbf{Detection}} & \multicolumn{2}{c|}{\textbf{re-ID}}                                  \\ \cline{2-5}
                                         & \textbf{Recall}                         & \multicolumn{1}{c|}{\textbf{AP}}    & \textbf{mAP}  & \textbf{top-1} \\ \hline \hline
        \multicolumn{1}{|c}{}            & \multicolumn{4}{c|}{\textbf{CUHK-SYSU}}                                                                        \\ \hline
        parallelization                  & 92.6                                    & 86.6                                & 91.7          & 92.8           \\
        serialization                    & 90.9                                    & 85.7                                & \textbf{92.5} & \textbf{93.7}  \\ \hline
        \multicolumn{1}{|c}{}            & \multicolumn{4}{c|}{\textbf{PRW}}                                                                              \\ \hline
        parallelization                  & 93.8                                    & 88.7                                & 43.6          & 80.0           \\
        serialization                    & 93.7                                    & 89.4                                & \textbf{44.7} & \textbf{80.8}  \\ \hline
    \end{tabular}
    }
    \caption{Influence of inferior features on the performance on CUHK-SYSU and PRW datasets. We separate the person search into detection and re-ID, and evaluate their performance individually. The bold font represents the best result. Most experiment results will be presented in this form.}
    \label{feature_misalignment}
\end{table}
As aforementioned, the baseline suffers from the inferior features. To investigate its influence, we train the baseline model and report the results under two evaluation settings in Table \ref{feature_misalignment}. The original setting is denoted by \textit{parallelization}. In the second setting called \textit{serialization}, the network will iterate twice to solve the detection and re-ID in turn. The first iteration will output detected BBoxes. Then we exploit RoI-Align to pool a fixed size region for each BBox and feed them into res5 to extract re-ID features. In this way, superior BBoxes can be obtained before re-ID stage. Table \ref{feature_misalignment} shows the mAP of re-ID is increased by 0.75\% on CUHK-SYSU, 1.12\% on PRW. This demonstrates the re-ID ability of the network is greatly limited by the inferior features of proposals. We also notice that the detection performance has a slight decline. This is caused by the inconsistency between the training and test phase, \textit{i.e.}, the network is trained by the proposals generated by RPN, but tested by the detected BBoxes. Therefore it is necessary to introduce serialization into model training, rather than just in test phase.

\subsection{Proposed Sequential End-to-end Network}
The overview of our model is shown in Figure \ref{net_arch} (b). It consists of two head networks to solve person detection and person re-ID respectively. The first standard Faster R-CNN head is employed to generate accurate BBoxes. The second unmodified baseline head is applied to further fine-tune these BBoxes and extract their discriminative features.

The main idea is to exploit Faster R-CNN as a stronger RPN to provide fewer but more accurate candidate BBoxes. These high-quality BBoxes lead to more discriminative embeddings.

\subsubsection{Training}
During training phase, these two heads are trained with 0.5 IoU threshold to distinguish positive and negtive samples, and the feature learning is supervised by the following 5 losses.
\begin{itemize}
    \item $L_{reg_1}/L_{reg_2}$: The regression loss of the first/second head. $N_p$ is the number of positive samples, $r_i$ is the calculated regressor of i-th positive sample, $\triangle_i$ is the corresponding ground truth regressor, and $L_{loc}$ is the Smooth-L$_1$-Loss.
          \begin{equation}
              L_{reg}=\frac{1}{N_p}\sum_{i=1}^{N_p}L_{loc}(r_i,\triangle_i)
          \end{equation}
    \item $L_{cls_1}$: The classification loss of the first head. $N$ is the number of samples, $p_i$ is the predicted classification probability of i-th sample, and $c_i$ is the ground truth label.
          \begin{equation}
              L_{cls_1}=-\frac{1}{N}\sum_{i=1}^{N}c_ilog(p_i)
          \end{equation}
    \item $L_{cls_2},L_{reid}$: The classification and re-ID losses of the second head. It is calculated by the Norm-Aware-Embedding $L_{nae}(.)$. $f$ is the extracted 256-dim features.
          \begin{equation}
              L_{cls_2},L_{reid}=L_{nae}(f)
          \end{equation}
\end{itemize}

The overall learning objective function is given as:
\begin{equation}
    L = \lambda_1L_{reg1} + \lambda_2L_{cls1} + \lambda_3L_{reg2} + \lambda_4L_{cls2} + \lambda_5L_{reid}
\end{equation}

$\lambda_1$ is set to 10, and the others are 1.

\subsubsection{Inference}
In inference stage, NMS is applied to remove redundant BBoxes before re-ID stage. In this way, the inference speed will be greatly accelerated.

\subsubsection{First Classification Score (FCS)}
\begin{figure}[t]
    \centering
    \resizebox{0.7\columnwidth}{!}
    {
        \begin{tikzpicture}
            \begin{axis}
                [ymin=0,
                    ymax=3000,
                    minor y tick num=3,
                    area style,
                    xlabel=IoU,
                    ylabel=Number of BBoxes]
                \addplot+[ybar interval,mark=no]
                plot coordinates {(0.5, 100) (0.6, 166) (0.7, 1000) (0.8, 2700) (0.9, 1600) (1.0, 0)};
            \end{axis}
        \end{tikzpicture}
    }
    \caption{The IoU statistics of the BBoxes of labeled pedestrians detected by the first head in test phase.}
    \label{iou_statistics}
\end{figure}
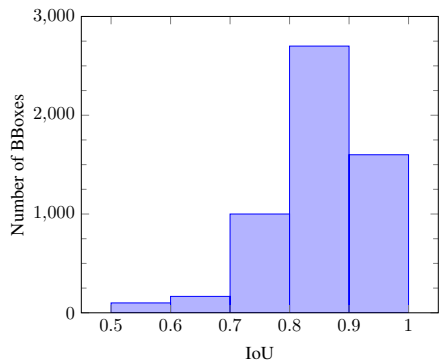
Figure \ref{iou_statistics} shows that there are a lot of detected BBoxes with IoU $>$ 0.8 in test phase. The second head is trained with 0.5 IoU threshold, so it may fail to classify these high IoU samples correctly. Hence we take the more reliable classification scores predicted by the first head as output.

\subsubsection{Discussion}
Our SeqNet shares the similar structure with previous works \cite{iterative-bbox1,iterative-bbox2,cascade-rcnn}, but our method differs from them significantly from the following aspects:

\begin{itemize}
    \item \textbf{Motivation} Previous multi-stage Faster R-CNN is proposed to achieve better detection performance. However, our method aims at solving the detection and re-ID sequentially with a jointly optimized network to extract more discriminative features.
    \item \textbf{Efficiency} Our SeqNet owns an extra NMS, which ensures the efficiency in test phase. In contrast, each head of their networks needs to handle all BBoxes.
\end{itemize}

\subsection{Context Bipartite Graph Matching}
In this section, we present a novel Context Bipartite Graph Matching (CBGM) algorithm used in test phase to integrate context information into the matching process.

Traditional person search task can be seen as a single-point matching strategy, which takes the person most similar to the querier in the gallery image as the search result. But it may fail when there are multiple people with very similar appearances in the gallery image. We extend it to a multi-point matching strategy, which matches both the querier and its surrounding people with all the detected pedestrians in the gallery image. In this way, when the single-point matching strategy fails, as long as the surrounding people can be correctly matched, the query person can still be identified.

Taking Figure \ref{matching_example} for example, we define the following symbols.
\begin{itemize}
    \item $Q/G$: The query/gallery image (the upper/lower image).
    \item $q$: The query person in $Q$ (the blue box, \textit{i.e.}, person (a)).
    \item $V$: All people in image ($V_G=\{(c),(d)\}$).
    \item $sim(p_1, p_2)$: The cosine similarity between person $p_1$ and $p_2$ calculated by extracted features.
    \item $SIM(q,G)$: The similarity between $q$ and $G$. It is defined as the maximum value among these similarities between $q$ and all people in $G$.
    \begin{equation}
        SIM(q,G)=\max_{p\in V_G}sim(q, p)
    \end{equation}
\end{itemize}

In graph theory, a \textit{matching} $M=(\mathbb{V},\mathbb{E})$ in an undirected graph is a set of edges without common vertices. $\mathbb{V}$ is the set of vertices and $\mathbb{E}$ is the set of edges. We further define the following concepts.
\begin{itemize}
    \item $weight(e_i,e_j)$: The weight of the edge $(e_i,e_j)\in \mathbb{E}$.
    \item $weight(M)$: The weight of matching $M$. It is defined as the sum of the weights of all edges.
    \begin{equation}
        weight(M)=\sum_{(e_i,e_j)\in \mathbb{E}}weight(e_i,e_j)
    \end{equation}
    \item $C(M)$: The confidence of matching $M$. It is defined as the maximum value among all weights.
    \begin{equation}
        C(M)=\max_{(e_i,e_j)\in \mathbb{E}}weight(e_i,e_j)
    \end{equation}
\end{itemize}

Based on these two sets of vertices $V_Q$ and $V_G$, we firstly build a complete bipartite graph $G=(\mathbb{V},\mathbb{E})$, in which $\mathbb{V}=V_Q\cup V_G$. The graph has the following properties:
\begin{itemize}
    \item For every two vertices $v_1\in V_Q$ and $v_2\in V_G$, $(v_1,v_2)$ is an edge in $\mathbb{E}$.
    \item No edge has both endpoints in the same set of vertices.
    \item For each edge $(e_i,e_j)\in \mathbb{E}$, its weight is the similarity of corresponding vertices, \textit{i.e.}, $weight(e_i,e_j)=sim(e_i,e_j)$.
\end{itemize}

Then the Kuhn-Munkres (K-M) algorithm \cite{km-k,km-m} is exploited to find the optimal matching with largest $weight(M)$. In Figure \ref{matching_example}, the matching is $(a)\leftrightarrow(c), (b)\leftrightarrow(d)$, and the query person (a) can be correctly matched with the ground truth (c).

The proposed Context Bipartite Graph Matching (CBGM) algorithm is described in Algorithm \ref{CBGM}. We rank all the gallery images in descending order by $SIM(q, G)$, and remain the top-$k_1$ to be processed. In this way, most gallery images in which $q$ does not appear can be removed. Additionally, excessive context information may bring noise. Therefore we only regard the people with top-$k_2$ detection confidence in the query image as context information. After the optimal matching $M$ is found, $C(M)$ is taken as the similarity between $q$ and its matched person.

\begin{algorithm}[t]
    \DontPrintSemicolon
      \KwInput
      {
          ~~\\
          Query image, $Q$\\
          Query person, $q\in V_Q$\\
          Gallery images, $S=\{G_1,G_2,...\}$\\
          Number of processed gallery images, $k_1$\\
          Maximum context, $k_2$
      }
      \KwOutput
      {
          ~~\\
          Most similar person in each gallery image\\
          Similarities between $q$ and these most similar people
      }
      Rank $S$ in descending order by $SIM(q,G)$\\
      Remain top-$k_1$ gallery images, $S=\{G_1,G_2,...,G_{k_1}\}$\\
      Rank $V_Q$ in descending order by detection confidence\\
      Remain top-$k_2$ people, $V_Q=\{q_1,q_2,...,q_{k_2}\}$\\
      Set $people,sims$ to empty list\\
      \For{each $G\in S$}
      {
          Based on $V_Q$ and $V_G$, build a complete bipartite graph $G=(\mathbb{V},\mathbb{E})$\\
          Exploit K-M algorithm to find the optimal matching $M$ with largest weight\\
          \For{each edge $(e_i,e_j)$ of M}
          {
              \If{$e_i=q$}
              {
                  Insert $e_j$ into $people$\\
                  Insert $C(M)$ into $sims$\\
                  break
              }
          }
      }
      return $people,sims$
    \caption{CBGM}
    \label{CBGM}
\end{algorithm}


\section{Experiments}
In this section, we first introduce the datasets and evaluation protocols. Then we describe the implementation details, followed by ablation studies on the efficacy of each component. Finally, we compare our method with state-of-the-art ones.

\subsection{Datasets and Evaluation Protocol}
\subsubsection{CUHK-SYSU}
CUHK-SYSU \cite{oim} is a large scale person search dataset containing 18,184 scene images and 96,143 annotated BBoxes, which are collected from two sources: street snap and movie. All people are divided into 8,432 labeled identities and other unknown ones. The training set contains 11,206 images and 5,532 different identities. The test set contains 6,978 images and 2,900 query people. The training and test sets have no overlap on images and query people. For each query, different gallery sizes from 50 to 4000 are pre-defined to evaluate the search performance. If not specify, gallery size of 100 is used by default.

\subsubsection{PRW}
PRW is another widely used dataset \cite{prw} containing 11,816 video frames captured by 6 cameras in Tsinghua university. 34,304 BBoxes are annotated manually. Similar to CUHK-SYSU, all people are divided into labeled and unlabeled identities. The training set contains 5,704 images and 482 different people, while the test set includes 6,112 images and 2,057 query people. For each query, the gallery is the whole test set, \textit{i.e.}, the gallery size is 6112.

\subsubsection{Evaluation Protocol}
Following the settings in previous works \cite{qeeps,nae}, the Cumulative Matching Characteristic (CMC) and the mean Averaged Precision (mAP) are adopted as the performance metrics. The formal is widely used in person re-ID, and the latter is inspired by object detection task. The higher the two metrics, the better the performance.

\subsection{Implementation Details}
We implement our model with PyTorch \cite{pytorch} and run all experiments on one NVIDIA Tesla V100 GPU. We adopt ResNet50 \cite{resnet} pretrained on the ImageNet \cite{imagenet} as the backbone network. During training, batch size is 5 and each image is resized to $900\times1500$ pixels. Our model is optimized by Stochastic Gradient Descent (SGD) for 20 epochs (18 epochs for PRW) with initial learning rate of 0.003 which is warmed up during the first epoch and decreased by 10 at the 16-th epoch. The momentum and weight decay of SGD are set to 0.9 and $5\times10^{-4}$ individually. For CUHK-SYSU/PRW, the circular queue size of OIM is set to 5000/500. At test time, NMS with 0.4/0.5 threshold is used to remove redundant boxes detected by the first/second head.

\subsection{Ablation Study}
In this section, we perform several analytical experiments on CUHK-SYSU to better understand our proposed method.

\subsubsection{Different detectors and re-identifiers}
We first explore whether the improvement brought by SeqNet comes from better detection or more discriminative features. We separate the person search task into two stages: detection stage with different detectors and re-ID stage with different re-identifiers. When using NAE re-identifier, we remove its RPN module and set the proposals manually to the BBoxes detected by the specified detector (\textit{e.g.} SeqNet). In particular, NAE detector + NAE re-identifier is equivalent to the \textit{serialization} mentioned in last section. The results are summarized in Table \ref{diff_detectors_identifiers}, from which we can draw the following conclusions:

\begin{table}[t]
    \centering
    \resizebox{0.8\columnwidth}{!}
    {
        \begin{tabular}{|l|cc|l|cc|}
            \hline
            \textbf{Detector}       & \textbf{Recall}       & \textbf{AP}           & \textbf{Re-identifier} & \textbf{mAP} & \textbf{top-1} \\ \hline \hline
            \multirow{2}{*}{NAE}    & \multirow{2}{*}{92.6} & \multirow{2}{*}{86.6} & NAE                     & 92.5         & 93.7           \\
                                    &                       &                       & SeqNet                  & 93.1         & 94.0           \\ \hline
            \multirow{2}{*}{SeqNet} & \multirow{2}{*}{92.1} & \multirow{2}{*}{89.2} & NAE                     & 93.3         & 93.8           \\
                                    &                       &                       & SeqNet                  & 93.8         & 94.5           \\ \hline
            \multirow{2}{*}{GT}     & \multirow{2}{*}{100}  & \multirow{2}{*}{100}  & NAE                     & 94.1         & 94.6           \\
                                    &                       &                       & SeqNet                  & 94.6         & 95.3           \\ \hline
        \end{tabular}
    }
    \caption{Analytical experiment results with different detectors and re-identifiers on CUHK-SYSU.}
    \label{diff_detectors_identifiers}
\end{table}

\begin{itemize}
    \item \textbf{The detection of SeqNet is better} We can see from the second column that SeqNet (Recall: 92.1, AP: 89.2) achieves better detection than NAE (Recall: 92.6, AP: 86.6) in overall. It is mainly because that each head (RPN head/Faster R-CNN head/baseline head) of SeqNet will perform regression to BBoxes, which makes our model more selective against false positives.
    \item \textbf{SeqNet is more discriminative for re-ID} When using NAE detector, the mAP and top-1 accuracy of SeqNet outperform that of NAE by 0.6\% and 0.3\% respectively. Similar improvement (mAP $\uparrow$ 0.5\%, top-1 $\uparrow$ 0.7\%) can be observed when using SeqNet detector. This demonstrates our SeqNet can extract more discriminative features with the same detection ability. This is caused by the inconsistency of NAE, \textit{i.e.}, trained by low-quality proposals but testd by high-quality detected BBoxes. In contrast, the baseline head of SeqNet is trained with detected BBoxes, which makes it more suitable for test scenario.
    \item \textbf{Detection is not the performance bottleneck} If ground truth BBoxes are adopted as detection results, the mAP of NAE can be increased by 2.4\%, while SeqNet can only be increased by 0.8\%. It indicates that SeqNet gains very little from better detection, and future research should focus on how to achieve a better re-ID.
\end{itemize}

\subsubsection{FCS and NMS}
SeqNet has two key components: FCS to improve the classification ability, NMS to accelerate the inference speed. The upper block of Table \ref{different_components} shows that FCS greatly improves the detection (Recall: 91.5$\rightarrow$92.7, AP: 86.7$\rightarrow$89.7), which leads to a better re-ID (mAP: 93.1$\rightarrow$93.8, top-1: 94.0$\rightarrow$94.5). In addition, although NMS slightly reduces detection performance (Recall: 92.7$\rightarrow$92.1, AP: 89.7→89.2), it does not affect re-ID and increases the FPS (processed Frames Per Second) from 7.4 to 11.5.

\subsubsection{The multi-task framework of baseline head}
The lower block of Table \ref{different_components} reports the impact of each task of the baseline head. Since the detection of Faster R-CNN head is strong enough, the regression and classification branches of baseline head will not have much impact on the overall detection, but they will facilitate the re-ID branch to learn more discriminative features. We can observe that neither regression nor classification branch alone can achieve the best performance, suggesting that the two are complementary.

\begin{table}[t]
    \centering
    \resizebox{\columnwidth}{!}
    {
        \begin{tabular}{|ccccc|cc|cc|c|}
            \hline
            \multirow{2}{*}{\textbf{FCS}} & \multirow{2}{*}{\textbf{NMS}} & \multirow{2}{*}{\textbf{Re-ID}} & \multirow{2}{*}{\textbf{Cls}} & \multirow{2}{*}{\textbf{Reg}} & \multicolumn{2}{c|}{\textbf{Detection}} & \multicolumn{2}{c|}{\textbf{Re-ID}} & \multirow{2}{*}{\textbf{FPS}}                         \\ \cline{6-9}
                                          &                               &                                 &                               &                               & \textbf{Recall}                         & \textbf{AP}                         & \textbf{mAP}                  & \textbf{top-1} &      \\ \hline \hline
                                          &                               & \checkmark                      & \checkmark                    & \checkmark                    & 91.5                                    & 86.7                                & 93.1                          & 94.0           & 7.4  \\
            \checkmark                    &                               & \checkmark                      & \checkmark                    & \checkmark                    & 92.7                                    & 89.7                                & 93.8                          & 94.5           & 7.4  \\
                                          & \checkmark                    & \checkmark                      & \checkmark                    & \checkmark                    & 89.1                                    & 86.3                                & 93.4                          & 94.4           & 11.5 \\
            \checkmark                    & \checkmark                    & \checkmark                      & \checkmark                    & \checkmark                    & 92.1                                    & 89.2                                & 93.8                          & 94.5           & 11.5 \\ \hline
            \checkmark                    & \checkmark                    & \checkmark                      &                               &                               & 92.6                                    & 89.5                                & 93.0                          & 93.8           & -    \\
            \checkmark                    & \checkmark                    & \checkmark                      &                               & \checkmark                    & 92.8                                    & 89.5                                & 93.3                          & 94.1           & -    \\
            \checkmark                    & \checkmark                    & \checkmark                      & \checkmark                    &                               & 92.4                                    & 89.4                                & 93.4                          & 94.2           & -    \\ \hline
        \end{tabular}
    }
    \caption{Influence of different components on accuracy and speed. The ablation study about FCS and NMS is in the upper block. The multi-task framework of baseline head is discussed in the lower block.}
    \label{different_components}
\end{table}


\subsubsection{Different $k_1$ and $k_2$ of CBGM}
\begin{table}[t]
    \centering
    \resizebox{0.6\columnwidth}{!}
    {
        \begin{tabular}{|c|ccccc|}
            \hline
            \diagbox{$\boldsymbol{k_2}$}{$\boldsymbol{k_1}$} & \textbf{10}                             & \textbf{20} & \textbf{30}   & \textbf{40} & \textbf{50} \\ \hline \hline
            \multicolumn{1}{|c}{}                            & \multicolumn{5}{c|}{\textbf{CUHK-SYSU}}                                                           \\ \hline
            \textbf{2}                                       & 94.9                                    & 94.8        & 94.8          & 94.8        & 94.8        \\
            \textbf{3}                                       & \textbf{95.2}                           & 95.1        & 95.0          & 95.0        & 95.0        \\
            \textbf{4}                                       & 95.1                                    & 94.9        & 94.7          & 94.6        & 94.6        \\
            \textbf{5}                                       & 95.0                                    & 94.7        & 94.6          & 94.5        & 94.4        \\
            \textbf{6}                                       & 95.0                                    & 94.8        & 94.7          & 94.5        & 94.5        \\ \hline
            \multicolumn{1}{|c}{}                            & \multicolumn{5}{c|}{\textbf{PRW}}                                                                 \\ \hline
            \textbf{2}                                       & 66.0                                    & 66.6        & 66.7          & 66.5        & 66.5        \\
            \textbf{3}                                       & 66.4                                    & 67.0        & 67.3          & 67.2        & 67.1        \\
            \textbf{4}                                       & 66.6                                    & 67.2        & \textbf{67.6} & 67.4        & 67.1        \\
            \textbf{5}                                       & 66.6                                    & 67.2        & 67.5          & 67.3        & 67.0        \\
            \textbf{6}                                       & 66.6                                    & 67.2        & 67.5          & 67.5        & 67.2        \\ \hline
        \end{tabular}
    }
    \caption{The performance on CUHK-SYSU (the upper block) and PRW (the lower block) datasets with different $k_1$ and $k_2$ of CBGM. We evaluate the performance by $\frac{mAP+top-1}{2}$.}
    \label{k1_k2}
\end{table}
We evaluate the performance with different $k_1$ and $k_2$ of CBGM. Figure \ref{k1_k2} shows that CBGM is robust to these two parameters. On CUHK-SYSU, $k_1/k_2=10/3$ achieves the best performance, while on PRW, $k_1/k_2=30/4$ is the best choice. This is because that the gallery size of PRW (6112) is much larger than that of CUHK-SYSU (100). Therefore, $k_1$ and $k_2$ needs to be larger to capture more context information for precise search.

\subsubsection{Efficiency of CBGM}
Table \ref{cbgm_speed} reports the average time to search a query person under different gallery sizes. For CUHK-SYSU dataset, after sorting all gallery images in descending order by $SIM(q, G)$, CBGM is applied to the top-10 gallery images. Table \ref{cbgm_speed} shows that the additional computation brought by CBGM is fixed (about 2ms) and light. The larger the gallery size, the smaller the impact of CBGM on speed.

\begin{table}[t]
    \centering
    \resizebox{0.8\columnwidth}{!}
    {
        \begin{tabular}{|l|ccccc|}
            \hline
            \multicolumn{1}{|c|}{\multirow{2}{*}{\textbf{Method}}} & \multicolumn{5}{c|}{\textbf{Gallery size}}                                       \\ \cline{2-6} 
            \multicolumn{1}{|l|}{}                                 & \textbf{100}  & \textbf{500}  & \textbf{1000}   & \textbf{2000}  & \textbf{4000} \\ \hline \hline
            SeqNet                                                 & 347           & 366           & 390             & 439            & 541           \\
            SeqNet+CBGM                                            & 349           & 368           & 392             & 441            & 542           \\ \hline
        \end{tabular}
    }
    \caption{The average time to search a query person under different gallery sizes on CUHK-SYSU. The unit is milliseconds.}
    \label{cbgm_speed}
\end{table}

\subsubsection{Integrated into another method}
To verify the universality of our method, we integrate SeqNet into the existing end-to-end framework. We choose the widely studied OIM \cite{oim} as the base network and its implementation is the same as in \cite{nae}. As shown in Table \ref{compare_with_sota}, SeqNet improves the mAP of OIM by 6.3\% and 11.8\% on CUHK-SYSU and PRW benchmarks respectively, which demonstrates that our method is insensitive to base network. Particularly, OIM+SeqNet+CBGM further achieves 94.3 of mAP and 95.0 of top-1 accuracy, outperforming all other competitors. It indicates the great potential of our method.

\subsection{Comparison with the State-of-the-art Methods}
In this section, we compare our method with the state-of-the-art models on CUHK-SYSU and PRW.

\begin{table}[t]
    \centering
    \huge
    \resizebox{\columnwidth}{!}
    {
        \begin{tabular}{|l|l|cc|cc|}
            \hline
            \multicolumn{2}{|c|}{\multirow{2}{*}{\textbf{Method}}} & \multicolumn{2}{c|}{\textbf{CUHK-SYSU}} & \multicolumn{2}{c|}{\textbf{PRW}}                                                  \\ \cline{3-6}
            \multicolumn{2}{|c|}{}                                 & \textbf{mAP}                            & \textbf{top-1}                    & \textbf{mAP}  & \textbf{top-1}                 \\ \hline \hline
            \multirow{6}{*}{\rotatebox{90}{two-stage}}             & DPM\cite{dpm}                           & -                                 & -             & 20.5           & 48.3          \\
                                                                   & MGTS\cite{mgts}                         & 83.0                              & 83.7          & 32.6           & 72.1          \\
                                                                   & CLSA\cite{clsa}                         & 87.2                              & 88.5          & 38.7           & 65.0          \\
                                                                   & RDLR\cite{reid-driven}                  & 93.0                              & 94.2          & 42.9           & 70.2          \\
                                                                   & IGPN\cite{igpn}                         & 90.3                              & 91.4          & 47.2           & 87.0          \\
                                                                   & TCTS\cite{tcts}                         & 93.9                              & 95.1          & 46.8           & 87.5          \\ \hline
            \multirow{15}{*}{\rotatebox{90}{end-to-end}}           & OIM\cite{oim}                           & 75.5                              & 78.7          & 21.3           & 49.9          \\
                                                                   & IAN\cite{ian}                           & 76.3                              & 80.1          & 23.0           & 61.9          \\
                                                                   & NPSM\cite{npsm}                         & 77.9                              & 81.2          & 24.2           & 53.1          \\
                                                                   & RCAA\cite{rcaa}                         & 79.3                              & 81.3          & -              & -             \\
                                                                   & CTXGraph\cite{context}                  & 84.1                              & 86.5          & 33.4           & 73.6          \\
                                                                   & QEEPS\cite{qeeps}                       & 88.9                              & 89.1          & 37.1           & 76.7          \\
                                                                   & HOIM\cite{hoim}                         & 89.7                              & 90.8          & 39.8           & 80.4          \\
                                                                   & BINet\cite{binet}                       & 90.0                              & 90.7          & 45.3           & 81.7          \\
                                                                   & NAE\cite{nae}                           & 91.5                              & 92.4          & 43.3           & 80.9          \\
                                                                   & NAE+\cite{nae}                          & 92.1                              & 92.9          & 44.0           & 81.1          \\ \cline{2-6}
                                                                   & \textit{OIM(ours)}                      & 87.1                              & 88.5          & 34.0           & 75.9          \\
                                                                   & \textit{OIM+SeqNet(ours)}               & 93.4                              & 94.1          & 45.8           & 81.7          \\
                                                                   & \textit{OIM+SeqNet+CBGM(ours)}          & 94.3                              & 95.0          & 46.6           & 84.9          \\
                                                                   & \textit{NAE+SeqNet(ours)}               & 93.8                              & 94.6          & 46.7           & 83.4          \\
                                                                   & \textit{NAE+SeqNet+CBGM(ours)}          & \textbf{94.8}                     & \textbf{95.7} & \textbf{47.6}  & \textbf{87.6} \\ \hline
        \end{tabular}
    }
    \caption{Comparison of mAP and top-1 accuracy with the state-of-the-art methods on CUHK-SYSU and PRW. Our models are shown in italics.}
    \label{compare_with_sota}
\end{table}

\subsubsection{CUHK-SYSU}
\begin{figure}[t]
    \centering
    \resizebox{0.8\columnwidth}{!}
    {
        \begin{tikzpicture}
            \begin{axis}
                [height=6.9cm,
                    width=8cm,
                    grid=major,
                    xmin=0, xmax=4000,
                    ymin=50, ymax=100,
                    minor y tick num=3,
                    xlabel=Gallery size,
                    ylabel=mAP,
                    legend pos=outer north east]
                \addplot[mark=square*,red]  
                plot coordinates {(100, 94.8) (500, 91.9) (1000, 90.2) (2000, 87.8) (4000, 85.3)};
                \addplot[mark=triangle*,green!50!black!50]  
                plot coordinates {(100, 92.1) (500, 87) (1000, 85) (2000, 82) (4000, 79)};
                \addplot[mark=*,blue]  
                plot coordinates {(100, 79.3) (500, 71) (1000, 65) (2000, 61) (4000, 57)};
                \addplot[mark=pentagon*,magenta]  
                plot coordinates {(100, 76.3) (500, 67) (1000, 63) (2000, 58) (4000, 55)};
                \addplot[mark=diamond*,orange]  
                plot coordinates {(100, 77.9) (500, 68) (1000, 64) (2000, 58) (4000, 54)};
                \addplot[mark=square*,purple]  
                plot coordinates {(100, 75.5) (500, 65) (1000, 61) (2000, 57) (4000, 52)};
                \addplot[mark=diamond*,cyan]  
                plot coordinates {(100, 90) (500, 84) (1000, 82) (2000, 78) (4000, 75)};
                \addplot[mark=pentagon*,purple]  
                plot coordinates {(100, 84.1) (500, 78) (1000, 74) (2000, 71) (4000, 66)};
                \addplot[mark=triangle*,orange,dashed]  
                plot coordinates {(100, 83) (500, 77) (1000, 73) (2000, 70) (4000, 67)};
                \addplot[mark=square*,cyan,dashed]  
                plot coordinates {(100, 76.3) (500, 67) (1000, 63) (2000, 58) (4000, 55)};
                \addplot[mark=triangle*,magenta,dashed]  
                plot coordinates {(100, 93) (500, 89) (1000, 88) (2000, 85) (4000, 82)};
                \addplot[mark=diamond*,green!50!black!50,dashed]  
                plot coordinates {(100, 93.9) (500, 90.2) (1000, 89) (2000, 87) (4000, 83.5)};
                \legend{NAE+SeqNet+CBGM\\NAE\\RCAA\\IAN\\NPSM\\OIM\\BINet\\CTXGraph\\MGTS\\CLSA\\RDLR\\TCTS\\}
            \end{axis}
        \end{tikzpicture}
    }
    \caption{The mAP under different gallery sizes. The dashed lines represent two-stage methods and the solid lines represent end-to-end ones.}
    \label{larger_gallery_size}
\end{figure}
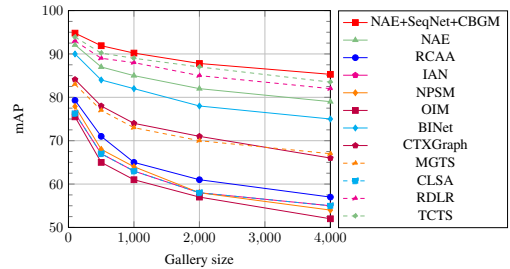
Table \ref{compare_with_sota} shows that both the mAP and top-1 accuracy of our method are higher than other competitors. Compared with the state-of-the-art two-stage model TCTS, our NAE+SeqNet+CBGM outperforms it by 0.9\% and 0.6\% w.r.t mAP and top-1 accuracy though it adopts more tricks, \textit{e.g.}, label smooth, random erasing \cite{random-erasing}, and triplet loss \cite{triplet-loss}.
This demonstrates the effectiveness of solving detection and re-ID jointly, which can avoid sub-optimal solution.

We also compare these methods under different gallery sizes. Figure \ref{larger_gallery_size} shows that the mAP of all algorithms decreases monotonically with the increase of gallery size, which indicates the difficulty of locating a target person in a large search scope. We can observe that our method still ranks best at all gallery sizes.

\subsubsection{PRW}
The right column of Table \ref{compare_with_sota} summarizes the results on PRW dataset. Our NAE+SeqNet+CBGM still surpasses the others. PRW has fewer training data than CUHK-SYSU, which indicates that our method is robust and effective for a relatively small dataset.

\subsubsection{Runtime Comparison}
\begin{table}[t]
    \centering
    \resizebox{0.9\columnwidth}{!}
    {
        \begin{tabular}{|l|ccccc|}
            \hline
            GPU(TFLOPs) & MGTS & QEEPS & NAE & NAE+ & SeqNet \\ \hline \hline
            K80(4.1)    & 1269 & -     & 663 & 606  & -      \\
            P6000(12.6) & -    & 300   & -   & -    & -      \\
            P40(11.8)   & -    & -     & 158 & 161  & -      \\
            V100(14.1)  & -    & -     & 83  & 98   & 86     \\ \hline
        \end{tabular}
    }
    \caption{Comparison of running time on different GPUs. The unit is milliseconds.}
    \label{time_comparison}
\end{table}
We compare the speed of different models, and show the Tera-Floating Point Operation per-second (TFLOPs) for each GPU for fair comparison. Our SeqNet is implemented in PyTorch, and images are resized to $900\times1500$ pixels, which is the same as MGTS and QEEPS. Table \ref{time_comparison} shows that our method is around 2 times faster than QEEPS and MGTS. Finally, our SeqNet costs 86 milliseconds per-frame on a V100 GPU, which is only a bit slower than NAE. The fast speed reveals the great potential of SeqNet to real-world applications.

\section{Conclusion}
In this paper, we notice the performance of previous end-to-end framework is limited by inferior features. To address the issue, we propose a Sequential End-to-end Network to solve the detection and re-ID in turn. Besides, we design a Context Bipartite Graph Matching algorithm to exploit context information as a complement to individual feature. Extensive experiments demonstrate that our method can significantly improve the performance of previous end-to-end models at an acceptable time cost.

\section{Acknowledgements}
This work is supported by National Key R\&D Program of China (Grant No. 213), the National Science Foundation of China (Grant No. 61976158 and 61673301).

\bibliography{refs/person_search.bib, refs/object_detection.bib, refs/person_reid.bib, refs/others.bib}

\end{document}